\documentclass[10pt,journal,compsoc]{IEEEtran}

\usepackage{graphicx}
\usepackage{amsmath,amssymb} 
\usepackage{color}
\usepackage{cite}
\usepackage{amsfonts}
\usepackage[utf8]{inputenc}
\usepackage{algpseudocode}
\usepackage{subfigure}
\usepackage{caption}
\usepackage{multirow}
\usepackage{fancyhdr}
\usepackage{amsthm}
\usepackage{textcomp}
\usepackage{amsmath}
\usepackage{mdwmath}
\usepackage{bm}
\usepackage{mathrsfs}
\usepackage{commath}
\usepackage{algorithm}
\usepackage[export]{adjustbox}
\usepackage{float}

\usepackage{esvect}

\begin{document}
\title{Automatic Recognition of Facial Displays of Unfelt Emotions}

\author{Kaustubh Kulkarni*, 
        Ciprian Adrian Corneanu*,
        Ikechukwu Ofodile*,~\IEEEmembership{Student Member,~IEEE,}
        Sergio Escalera,
        Xavier Bar\'o,
        Sylwia Hyniewska,~\IEEEmembership{Member,~IEEE,}
        J\"uri Allik,
        and~Gholamreza~Anbarjafari,~\IEEEmembership{Senior~Member,~IEEE}
\IEEEcompsocitemizethanks{\IEEEcompsocthanksitem I. Ofodile and G. Anbarjafari are  with the the iCV Research Group, Institute of Technology, University of Tartu, Tartu, Estonia.\protect\\
E-mail: \{ike,shb\}@icv.tuit.ut.ee

\IEEEcompsocthanksitem K. Kulkarni is with the Computer Vision Center, Barcelona, Spain.\protect\\
E-mail: kaustubh14jr@gmail.com

\IEEEcompsocthanksitem Ciprian A. Corneanu and S. Escalera are with the Computer Vision Center , University of Barcelona, Barcelona and University of Autonoma, Barcelona, Spain.\protect\\
E-mail: \{kaustubh14jr,cipriancorneanu\}@gmail.com, sergio@maia.ub.es
\IEEEcompsocthanksitem X. Bar\'o is with the Computer Vision Center and Universitat Oberta de Catalunya, Barcelona, Spain.\protect\\  Email: xbaro@uoc.edu
\IEEEcompsocthanksitem S. Hyniewska is with the Institute of Physiology and Pathology of Hearing, Poland.\protect\\
E-mail: s.hyniewska@bath.ac.uk
\IEEEcompsocthanksitem J. Allik is with Department of Psychology, University of Tartu and The Estonian Center of Behavioral and Health Sciences, Tartu, Estonia.\protect\\
E-mail: juri.allik@ut.ee
\IEEEcompsocthanksitem G.~Anbarjafari is also with Department of Electrical and Electronic Engineering, Hasan Kalyoncu University, Gaziantep, Turkey.
\IEEEcompsocthanksitem * Authors contributed equally in this work.}
\thanks{Manuscript received July 13, 2017; revised Xxxxx XX, 2017.}}

\markboth{Journal of IEEE Transactions on Affective Computing,~Vol.~XX, No.~X, Xxx~2017}%
{Ofodile \MakeLowercase{\textit{et al.}}: Fake Emotions}

\IEEEtitleabstractindextext{%
\begin{abstract}
Humans modify their facial expressions in order to communicate their internal states and sometimes to mislead observers regarding their true emotional states. Evidence in experimental psychology shows that discriminative facial responses are short and subtle. This suggests that such behavior would be easier to distinguish when captured in high resolution at an increased frame rate. We are proposing SASE-FE, the first dataset of facial expressions that are either congruent or incongruent with underlying emotion states. We show that overall the problem of recognizing whether facial movements are expressions of authentic emotions or not can be successfully addressed by learning spatio-temporal representations of the data. For this purpose, we propose a method that aggregates features along fiducial trajectories in a deeply learnt space. Performance of the proposed model shows that on average it is easier to distinguish among genuine facial expressions of emotion than among unfelt facial expressions of emotion and that certain emotion pairs such as contempt and disgust are more difficult to distinguish than the rest. Furthermore, the proposed methodology improves state of the art results on CK+ and OULU-CASIA datasets for video emotion recognition, and achieves competitive results when classifying facial action units on BP4D datase.

\end{abstract}

}

\maketitle

\IEEEdisplaynontitleabstractindextext

\IEEEpeerreviewmaketitle

\section{Introduction}
\label{sec:introduction}

\IEEEPARstart{I}n \emph{"Lie to me"}, an American crime television drama, Dr. Cal Lightman, a genius scientist, is assisting investigators in the police departments to solve cases through his knowledge of applied psychology. This is mainly done through interpreting subtle facial expressions of emotion (FEE) and body language of alleged offenders in order to evaluate their authentic motivation or emotional experience.\\
\begin{figure}
    \centering
    \includegraphics[width=0.7\linewidth]{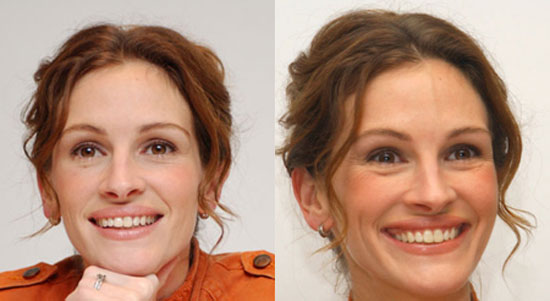}
    \caption{People may have difficulties in expressing emotions that look genuine when these do not correspond to the emotional state they are experiencing. In the case of smiling, differences can be observed in the contraction of the \emph{orbicularis oculi} muscle around the eyes. \textit{Left}: The lack of orbicularis oculi contraction has often been considered a marker of unfelt or even deceitful expressions. \textit{Right}: A strong orbicularis oculi contraction, with very visible "crows feet" around the corners of the eyes, has often been considered a marker of genuine expressions.}
    \label{fig:duchenne_smile}
\end{figure}
However in real life, humans are very skilled in concealing their true affective states from others and displaying emotional expressions that are appropriate for a given social situation. Untrained observers tend to perform barely above chance level when asked to detect whether observed behaviours genuinely reflect underlying emotions \cite{hartwig2011lie,ten2015physically}. This is a particularly difficult judgement when relying on visual cues only \cite{bond2006accuracy}. Even for professional psychologists it is difficult to recognise deceit in emotional displays as there are numerous factors that need to be considered \cite{porter2008reading,ochs2005intelligent}. \\
Many potential applications would benefit from the ability of automatically discriminating between subtle facial expressions such as displays of genuine and unfelt emotional states. Improved human-computer interaction, improved  human-robot interaction for assistive robotics \cite{bruce2002role,shibata1997artificial,lee2006can,anderson06}, treatment of chronic disorders \cite{littlewort2007faces} and assisting investigation conducted by police forces \cite{aremu2009path,vrij2001killed,o2009police} would be just a few. \\
An emotional display is considered unfelt (or masked) when it does not match a corresponding emotional state. There are three major ways in which emotional facial expressions are intentionally manipulated \cite{ekman1975unmasking}: an expression is \emph{simulated} when it is not accompanied by any genuine emotion, \emph{masked} when the expression corresponding to the felt emotion is replaced by a falsified expression that corresponds to a different emotion, or \emph{neutralized} when the expression of a true emotion is inhibited while the face remains neutral. All along this work, the term genuine FEE is used to denote FEEs congruent with the affective state, while the term unfelt FEE is used for denoting FEEs incongruent with the emotional state (aka masked) .\\
It has been argued that liers, deceivers and displayers of unfelt emotions would be betrayed by the leakage of their genuine emotional states through their nonverbal behaviour \cite{frank1997ability,abe2009neurobiology,porter2008reading}. This is supposed to happen through subtle facial expressions of short duration, as well as changes in pitch, posture and body movement. \\
Studies on the unfelt display of emotion mostly originated based on Duchenne de Boulogne's work, a nineteenth century French scientist. He is considered the first to have differentiated facial actions observed in displays of felt and unfelt emotions \cite{duchenne1862mechanism,spence2001behavioural}. Part of his legacy concerns what is considered the typical genuine smile -- often called a Duchenne smile. Duchenne smiles involve the contraction of the orbicularis oculi muscle (causing lifting of the cheeks and crow's feet around the eyes) together with the zygomaticus major muscle (pulling of lip corners upwards) \cite{bernstein2008adaptive,ekman1990duchenne,brown2002smile,frank1993not,ekman1988smiles,gunnery2013deliberate,krumhuber2009can,mehu2012reliable} (see Fig. \ref{fig:duchenne_smile}). In contrast, a masking smile (aka a non-Duchenne smile) can be used to conceal the experience of negative emotions \cite{frank2002smiles,ekman1988smiles,darwin1872expression,gosselin2010voluntary,mehu2012reliable}. \\
Although it has been argued that the orbicularis oculi activation is absent from masked facial expressions of enjoyment, empirical evidence is not conclusive. For example, in a database presenting 105 posed smiles  67\% of them were accompanied by the orbicularis oculi activation \cite{kanade2000comprehensive}. Another study showed that over 70\% of untrained participants were able to activate the majority of eye region action units, although not one action at a time, as they managed to perform them through the reliance and co-activation of other action units. The poorest performance was for the deliberate activation of the \emph{nasolabial furrow deepener}, which is often observed in sadness and which was performed successfully only by 20\% while the orbiculari oculi by 60\% of participants.\\
Although a variety of studies have focused on the evaluation of how genuine some FEEs might be while relying on the analysis of still, i.e. static, images, not much attention has been paid to dynamics as evaluated in a sequence of frames \cite{boraston2008brief,manera2011individual,uusberg2013unintentionality,perron2013analysis,chartrand2005judgement,vrij2010pitfalls,qu2017cas}. In a naturalistic setting, FEEs are always perceived as dynamic facial displays, and it is easier for humans to recognize facial behaviour in video sequences rather than in still images \cite{sato2004enhanced,krumhuber2013effects,jack2015human}. \\ 
It has been asserted that while trying to simulate the expression of an unfelt emotion, cues of the actual felt emotion appeared along cues related to the masked expression, which made the overall pattern difficult to analyse \cite{iwasaki2016hiding}. Leakages of a genuine emotion have been observed more frequently in the upper part of the face, while cues the lower half of the face was more often manipulated in order to express an unfelt emotion \cite{ross2013decoding,porter2012secrets,lusi2017joint,loob2017dominant}.\\
In this work, we propose a new data corpus containing genuine and unfelt FEE. While numerous studies involving the analysis of genuine or truthful behaviours rely on video recordings of directed interviews, such as the work in \cite{ten2015physically}, studies that analysed nonverbal behaviour while controlling for the emotional state of subjects are rare \cite{porter2012secrets}.\\
When designing experiments that require facial emotion displays as independent variables, posed facial expressions of subjects being instructed to act out a particular emotion are often used. This is thought to provide greater control over the stimuli than a spontaneous emotion display might, in the sense that other variables such as context and the physical appearance of subjects (even hair style or make-up) are much less variable and will not bias the observers in an uncontrolled way.\\
To record FEEs, participants are usually asked to practice the display of specific emotions. In order to achieve a display close to a genuine emotional expression, the process can be facilitated through the presentation of FEEs \cite{ekman2002facs,ekman1993facial}, or other pictures \cite{porter2012secrets} or videos inducing emotions in line with the ones to be expressed \cite{zhang2013high}, or mental imagery and related theatre techniques \cite{banziger2012introducing}. Such paradigms have been frequently used for recording and creating emotional expression databases \cite{gaebel1992facial,de2009rapid,calder2000configural,ekman1993facial,sandbach2012static,mavadati2013disfa, banziger2012introducing}.\\
In addition to the published dataset, we propose a complete methodology that has the capacity to recognise unfelt FEEs and generalises to standard public emotion recognition datasets. We first train a Convolutional Neural Network (CNN) to learn a static representation from still images and then pull features from this representation space along facial landmark trajectories. From these landmark trajectories we build final features from sequences of varying length using a Fisher Vector encoding which we use to train a SVM for final classification. State-of-the-art results are presented on CK+ and Oulu-Casia, two datasets containing posed FEEs. Moreover, close to state-of-the-art results are shown on a more difficult problem of recognising spontaneous facial Action Units on BP4D-Spontaneous. We finally provide benchmarking and outperform the methods from the recent ChaLearn Challenge \cite{wan2017results} on the proposed SASE-FE dataset.\\
The rest of the paper is organised as follows: in Section \ref{sec:related_work} we describe related work in FEEs recognition, in Section \ref{sec:dataset} we introduce the new SASE-FE dataset, in Section \ref{sec:method} we detail the proposed methodology, and Section \ref{sec:results} concludes the paper. 

\section{Related Work}
\label{sec:related_work}

This section first reviews main works on recognition of FEE, and then recognition of genuine and unleft FEE.

\subsection{Recognizing Facial Expressions of Emotion}
Automatic facial expression recognition (AFER) has been an active field of research for a long time. In general, a facial expression recognition system consists of four main steps. First the face is localised and extracted from the background. Then, facial geometry is estimated. Based on it, alignment methods can be used to reduce variance of local and global descriptors to rigid and non-rigid variations. Finally, representations of the face are computed either globally, where global features extract information from the whole facial region, or locally, and models are trained for classification or regression problems.\\
Features can be split into static and dynamic, with static features describing a single frame or image and dynamic ones including temporal information. Predesigned features can also be divided into appearance and geometrical. Appearance features use the intensity information of the image, while geometrical ones measure distances, deformations, curvatures and other geometric properties. This is not the case for learned features, for which the nature of the extracted information is usually unknown.\\
Geometric features describe faces through distances and shapes. These can be distances between fiducial points \cite{pantic2006dynamics} or deformation parameters of a mesh model \cite{sebe07, kotsia07}. In the dynamic case the goal is to describe how the face geometry changes over time. Facial motions are estimated from color or intensity information, usually through Optical flow \cite{wollmer2013lstm}. Other descriptors such as Motion History Images (MHI) and Free-Form Deformations (FFDs) are also used \cite{koelstra10}. Although geometrical features are effective for describing facial expressions, they fail to detect subtler characteristics like wrinkles, furrows or skin texture changes. Appearance features are more stable to noise, allowing for the detection of a more complete set of facial expressions, being particularly important for detecting micro-expressions.\\
Global appearance features are based on standard feature descriptors extracted on the whole facial region. Usually these descriptors are applied either over the whole facial patch or at each cell of a grid. Some examples include Gabor filters \cite{littlewort2011computer}, Local Binary Pattern (LBP) \cite{savran2014temporal,anbarjafari2013face}, Pyramids of Histograms of Gradients (PHOG) \cite{dhall2011emotion} and Multi-Scale Dense SIFT (MSDF) \cite{sun2014combining}.
Learned features are usually trained through a joint feature learning and classification pipeline. The resulting features usually cannot be classified as local or global. For instance, in the case of Convolutional Neural Networks (CNN), multiple convolution and pooling layers may lead to higher-level features comprising the whole face, or to a pool of local features. This may happen implicitly, due to the complexity of the problem, or by design, due to the topology of the network. In other cases, this locality may be hand-crafted by restricting the input data. \\
Expression recognition methods can also be grouped into static and dynamic. Static models evaluate each frame independently, using classification techniques such as Bayesian Network Classifiers (BNC) \cite{sebe07, cohen03learning}, Neural Networks (NN) \cite{tian01}, Support Vector Machines (SVM) \cite{kotsia07} and Random Forests (RF) \cite{dapogny2015dynamic}.
More recently, deep learning architectures have been used to jointly perform feature extraction and recognition. These approaches often use pre-training \cite{hinton06}, an unsupervised layer-wise training step that allows for much larger, unlabelled datasets to be used. CNNs are by far the dominant approach \cite{rifai12,liu2014learning,song2014deep}. It is a common approach to make use of domain knowledge for building specific CNN architectures for facial expression recognition. For example, in AU-aware Deep Networks \cite{LiuAURF}, a common convolutional plus pooling step extracts an over-complete representation of expression features, from which receptive fields map the relevant features for each expression. Each receptive field is fed to a DBN to obtain a non-linear feature representation, using an SVM to detect each expression independently. In \cite{LiuDBN} a two-step iterative process is used to train Boosted DBN (BDBN) where each DBN learns a non-linear feature from a face patch, jointly performing feature learning, selection and classifier training.\\
Dynamic models take into account features extracted independently from each frame to model the evolution of the expression over time. Probabilistic Graphical Models, such as Hidden Markov Models (HMM) \cite{koelstra10,le11,wu2015multi}, are common. Other techniques use Recurrent Neural Network (RNN) architectures, such as Long Short Term Memory (LSTM) networks \cite{wollmer2013lstm}. Some approaches classify each frame independently (e.g. with SVM classifiers \cite{geetha2009facial}), using the prediction averages to determine the final facial expression.
Intermediate approaches are also proposed where motion features between contiguous frames are extracted from interest regions, afterwards using static classification techniques \cite{sebe07}. For example, statistical information can be encoded at the frame-level into Riemannian manifolds \cite{liu2014combining}.\\

\subsection{Recognizing Genuine and Unfelt Facial Expressions of Emotion}
Emotion perception by humans or machines stands for the interpretation of particular representations of personal feelings and affects expressed by individuals, which may take different forms based on the circumstances governing their behaviour at the time-stamp at which they are evaluated \cite{diener2003personality,lucey2011automatically}.\\
Amongst audiovisual sources of information bearing clues to the emotions being expressed, the ones extracted from single or multiple samples of facial configurations, i.e. facial expressions, provide the most reliable basis for devising the set of criteria to be incorporated into the foregoing analysis \cite{iwasaki2016hiding,zhang2007real} and are, therefore, the most popular alternatives utilised in numerous contexts, such as forensic investigation and security.
These settings often rely on the assessment of the correspondence of the displayed expression to the actual one.

\section{SASE-FE Dataset}
\label{sec:dataset}
\begin{figure}[htp]
\centering
    \subfigure[Anger]
    {
        \includegraphics[width=0.21\textwidth]{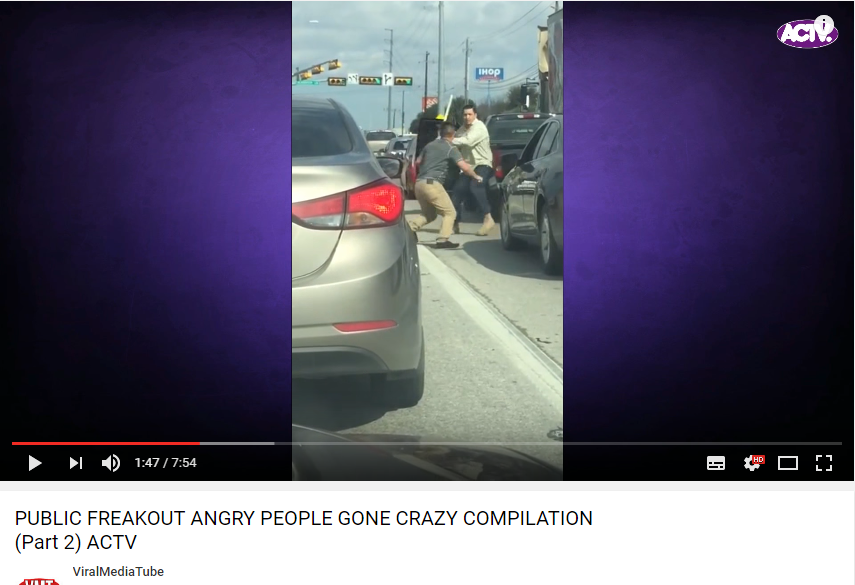}
    }   
    \subfigure[Happiness]
    {
        \includegraphics[width=0.21\textwidth]{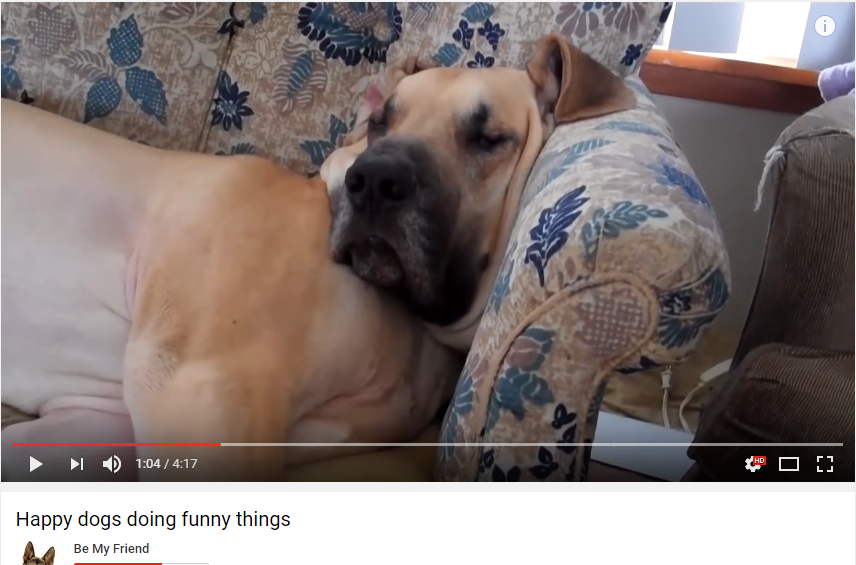}
    }
    \subfigure[Disgust]
    {
        \includegraphics[width=0.21\textwidth]{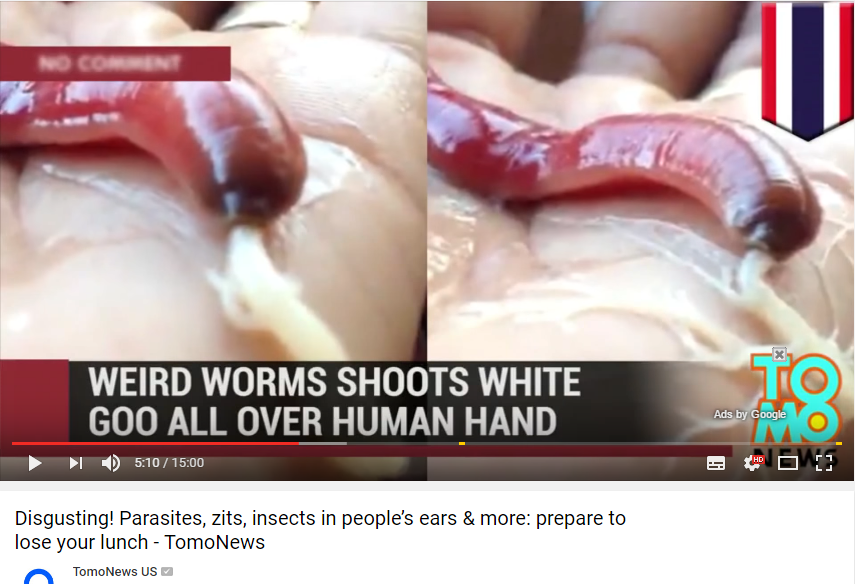}
    }
    \subfigure[Sadness]
    {
        \includegraphics[width=0.21\textwidth]{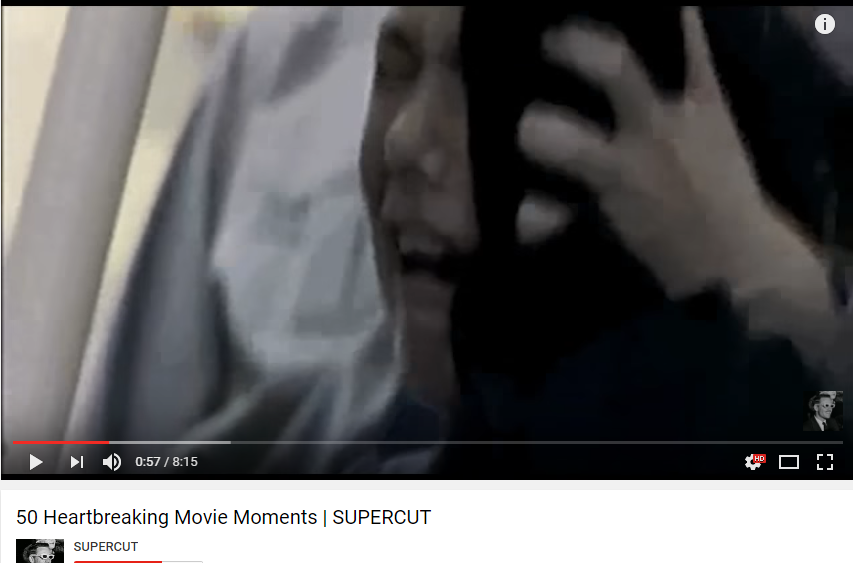}
    }
    \subfigure[Surprise]
    {
        \includegraphics[width=0.21\textwidth]{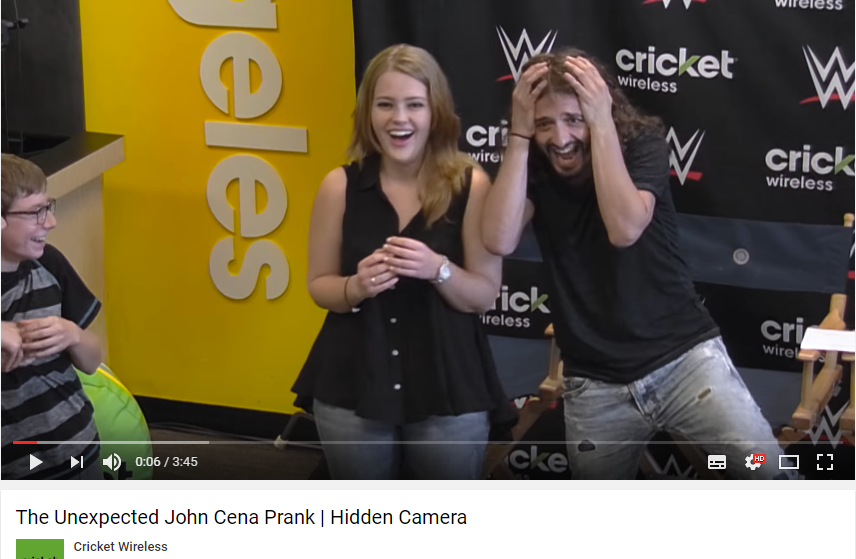}
    }
    \subfigure[Contempt]
    {
        \includegraphics[width=0.21\textwidth]{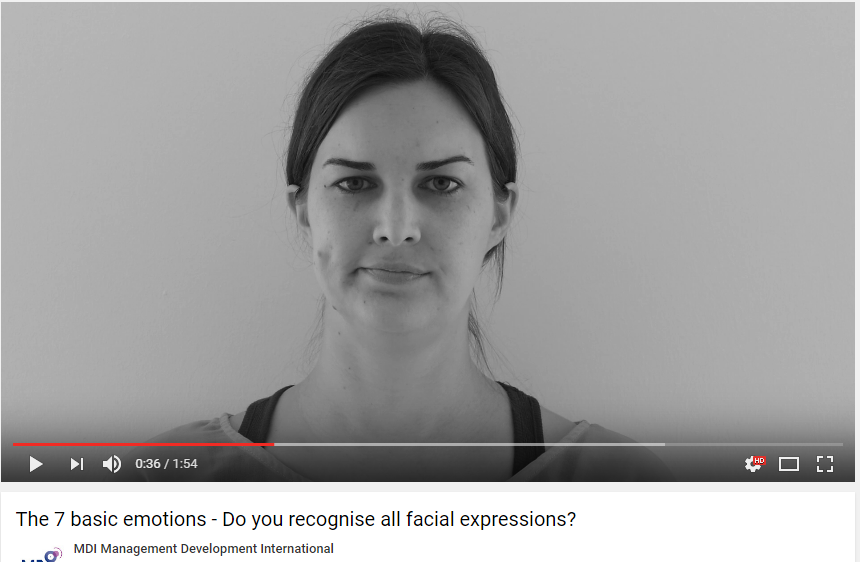}
    }
    \caption{A screenshot of some of the videos that have been used to induce a specific basic emotion in participants. }
    \label{vidEm}
\end{figure}

A number of affective portrayal databases exist; however, none meets the required criteria for our analysis of controlled genuine and unfelt emotional displays presented in high resolution at an increased frame rate. To answer those needs, the SASE-FE database was created.\\
The SASE-FE database consists of 643 different videos which had been recorded with a high resolution GoPro-Hero camera. From the inital 648 recordings, 5 were eliminated post-hoc as the participants did not completely meet the defined protocol criteria. As indicated in Table \ref{SASE-FE}, 54 participants of ages 19-36 were recorded. The reasoning behind the choice of such a young sample is that older adults have different, more positive responses than young adults about feelings and they are quicker to regulate negative emotional states than younger adults \cite{ready2016judgment,isaacowitz2012mood}. \\
Participants signed a written informed consent form after the experimental and recording procedures were explained. All participants agreed for their data to be released for research purposes and all data can be accessed by contacting the authors. The data collection and its use are based by the ethical rules stated by University of Tartu, Estonia.\\
For each recording, participants were asked to act two FEEs in a sequence, a genuine and an unfelt one. The participants displayed six universal expressions: Happiness, Sadness, Anger, Disgust, Contempt and Surprise. The subjects were asked if they felt the emotion and the large majority confirmed, but no recording of their answer was made.
To increase the chances of distinguishing between the two FEEs presented in a sequence, two emotions were chosen based on their visual and conceptual differences as observed on the two dimensions of valence and arousal \cite{plutchik1970emotions,jaimes2007multimodal,noroozi2017audio,larsen2011further}. Thus a visual contrast was created by asking participants to act Happy after being Sad, Surprised after being Sad, Disgusted after being Happy, Sad after being Happy, Angry after being Happy, and Contemptuous after being Happy \cite{whitesell1989children,mathieu2005annotation}. For eliciting emotion, subjects were shown videos in line with the target emotion. Emotion elicitation through videos is a well established process in emotion science research \cite{gross1995emotion}. Videos were short scenes from YouTube selected by psychologists .
Fig. \ref{vidEm} shows captures from videos that have been used for inducing specific emotions in the participants.\\

Throughout the entire setup, participants were asked to start their portrayals from the neutral face. The length of facial expression was about 3-4 seconds. After each genuine FEE, participants were asked to display a neutral state again and then the expression of a second emotion, which was the opposite of the former.\\
None of the participants were aware of the fact that they would be asked to display a second facial expression. The participant's first two seconds of behavior when performing a facial expression, and more exactly the opposite to the felt emotion, were recorded with the same device and the same configuration. As a result, for each participant we have collected 12 different videos of which 6 are genuine FEE and other 6 are unfelt FEE. The length of captured FEE is not fixed. The process has been closely supervised by experimental psychologists so that the setup would result in realistic recordings of genuine and unfelt FEE. The summary of the SASE-FE dataset is provided in Table \ref{SASE-FE}.\\
It is important to note that while preparing the SASE-FE database, introduced and used in this work, external factors such as personality or mood of the participants have been ignored, due to the fact that in order to eliminate such external factors several repetitions of the experiment would be necessary, but as a result the participant could start to learn to simulate the facial expressions better. Hence we have decided to ignore such external factors. 
\begin{table}
\centering
\caption{Summary of SASE-FE database.}
\begin{tabular}{|c|c|c|}
\hline
\multirow{4}{*}{\textbf{Subjects}} & \# of persons & 54  \\ \cline{2-3}
& gender distribution & female 41\%, male 59\% \\ \cline{2-3}
& age distribution & 19 - 36 years  \\ \cline{2-3}
& race distribution & \parbox[t]{3.5cm}{Caucasian 77.8\%, Asian 14.8\%, African 7.4\%} \\ \hline \hline
\multirow{4}{*}{\textbf{Videos}} & \# of videos & 643 \\ \cline{2-3} 
 & video length & 3-4 sec \\ \cline{2-3} 
 & resolution & 1280 $\times$ 960 \\ \cline{2-3} 
 & \#frames (acted/unfelt) & 120,216/118,712 \\ \hline
\end{tabular}
\label{SASE-FE}
\end{table}

\begin{figure}[htp]
    \centering
    \subfigure[Anger]
    {
        \includegraphics[width=0.40\textwidth]{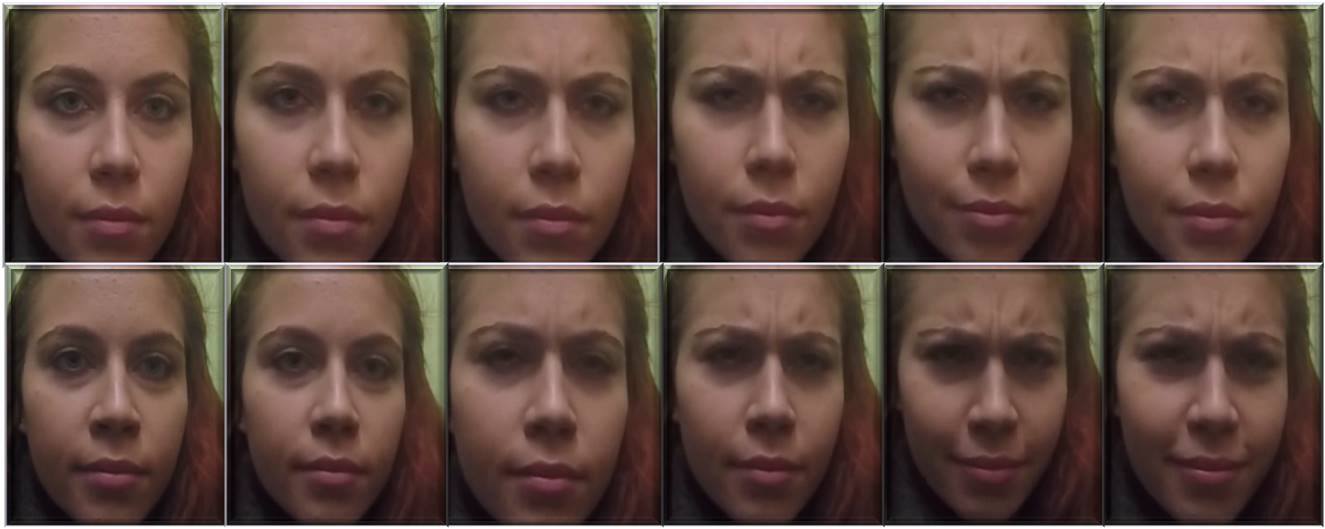}
        \label{fig:anger}
    }\quad    
    \subfigure[Happiness]
    {
        \includegraphics[width=0.40\textwidth]{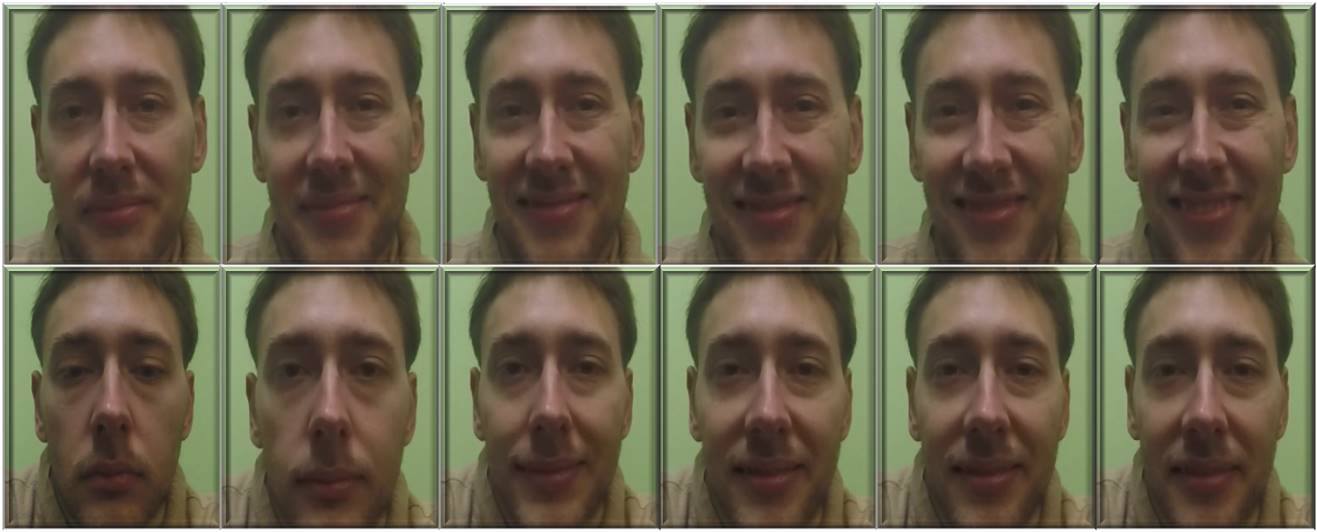}
        \label{fig:happiness}
    }\quad

    \subfigure[Surprise]
    {
        \includegraphics[width=0.40\textwidth]{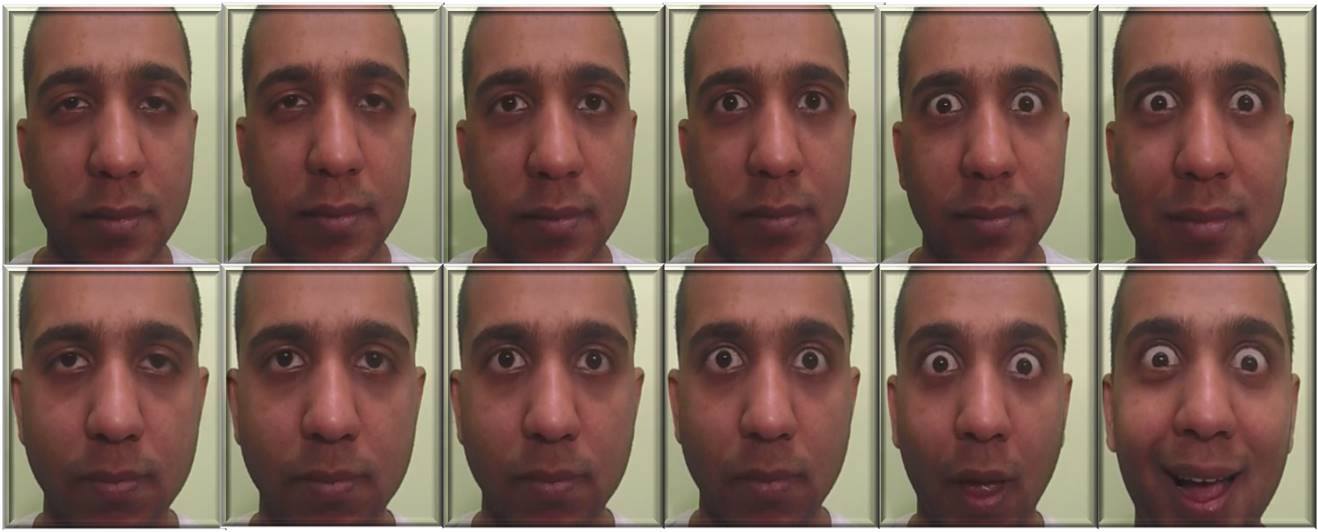}
        \label{fig:surprised}
    }\quad

    \caption{Selected examples of pairs of sequences showing genuine (top) and unfelt (below) FEEs of Anger, Happiness and Suprise from the SASE-FE dataset.}
    \label{fig:SASE-FE}
\end{figure}

\section{The Proposed Method}  
\label{sec:method}
\begin{figure*}
    \centering
    \includegraphics[width=0.8\linewidth, height=0.47\textwidth]{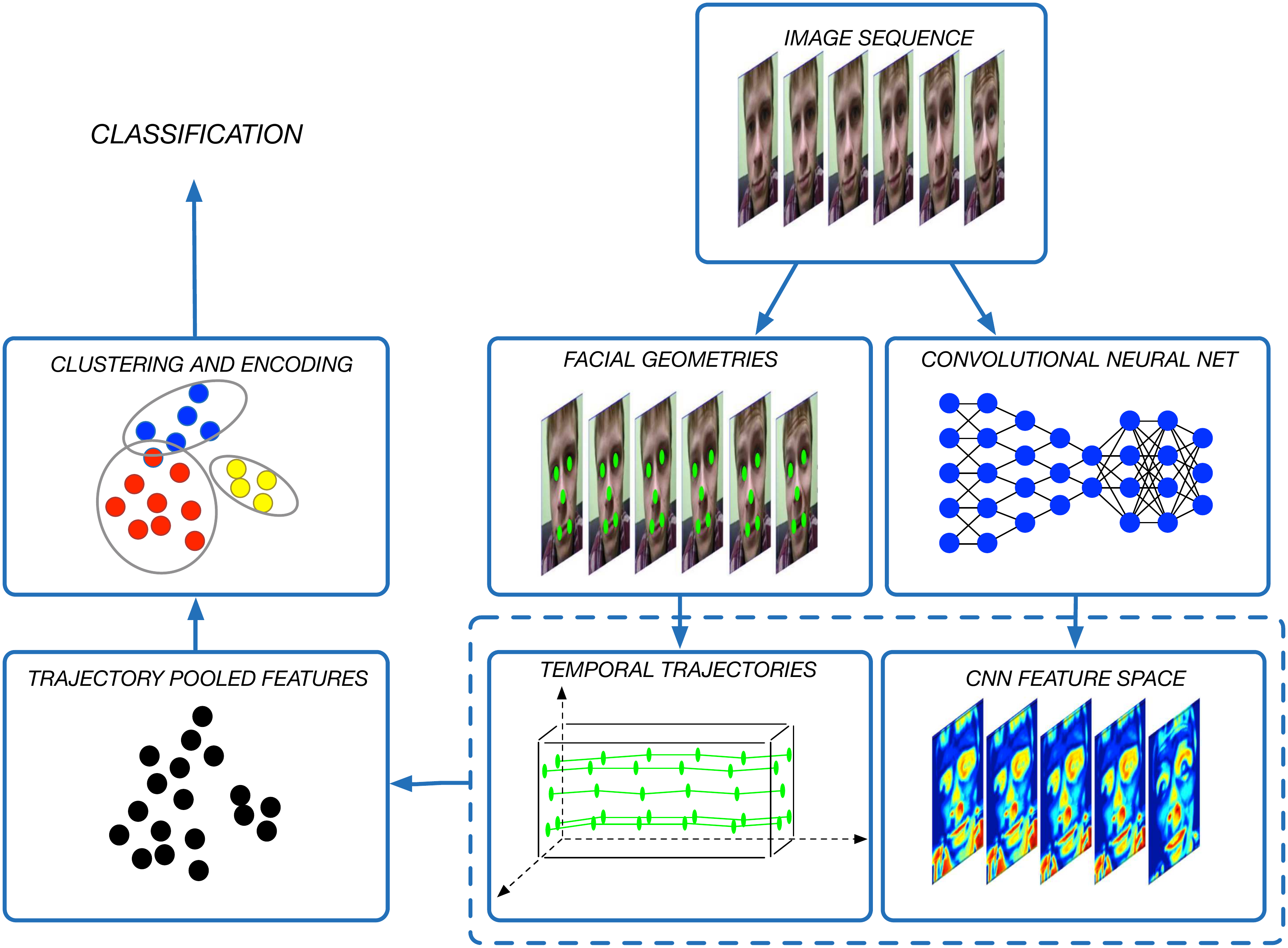}
    \caption{Overview of the proposed method.}
    \label{fig:method}
\end{figure*}

In this section, we present the methodology used for recognising unfelt FEEs from video sequences. As showed in the literature (see Sec. \ref{sec:introduction} and Sec. \ref{sec:related_work}) most discriminative information is to be found in the dynamics of such FEEs. Following this assumption, we consider learning a discriminative spatio-temporal representation to be central for this problem. We first train a Convolutional Neural Network (CNN) to learn a static representation from still images and then pull features from this representation space along facial landmark trajectories. From these landmark trajectories and inspired by previous work in action recognition \cite{wang2015action}, a well studied sequence modelling problem, we build final features from sequences of varying length using a Fisher Vector encoding which we use to train a SVM for final classification.\\
Additionally, the amount of video data available is limited, which requires usage of advanced techniques when training high capacity models with millions of parameters such as CNNs. Fine-tuning existing deep architectures can alleviate this problem to a certain extent but these models might carry redundant information from the pre-trained application domain. In this paper, we use a recently proposed method \cite{ding2017facenet2expnet} which proposes a regularisation function which helps using the face information to train the expression classification net.\\
We follow this section by first discussing the technique we have used to train a CNN on still images with a limited amount of data in Sec. \ref{sec:knowledge_transfer}. Then we show how we build a spatio-temporal representation from static features computed by the CNN in Sec. \ref{sec:spatio_temporal}. The reader can refer to Fig. \ref{fig:method} for an overview of the proposed method. Specific implementation details will be presented in Sec. \ref{sec:implementation}.

\subsection{Using efficient knowledge transfer for training a CNN for facial expression recognition}
\label{sec:knowledge_transfer}

Our proposed training procedure of the CNN for learning static spatial representation: first, we fine tune the VGG-Face network for the facial expression recognition task \cite{Parkhi15}. We then use this fine tuned network to guide the learning of a so called emotion network (EMNet) \cite{ding2017facenet2expnet}. Following \cite{ding2017facenet2expnet} the EMNet is denoted as:

\begin{equation}
 O = h_{\theta_2}(g_{\theta_1}(I))\:,
\end{equation}

where $h$ represents the fully connected layers and $g$ represents the convolution layers, $\theta_2$ and $\theta_1$ are the corresponding parameters of the to be estimated of the fully connected layers and the convolution layers respectively, $I$ is the input image and $O$ is the output before the softmax.\\
We follow the two step training proposed in \cite{ding2017facenet2expnet}. The basic motivation behind this training procedure is that the fine tuned VGG-Face network already gives a competitive performance on the emotion recognition task. We use the ouyput of the VGG-Face to guide the training of the EMNet. In the first step, we estimate the parameters of the  only of the convolution layers of the EMNet. In this step, the output of the VGG-Face acts as a regularisation for the emotion net. This step is achieved by maximising the following loss function:  

\begin{equation}
L_1 = \max_{\theta_1} {\lVert g_{\theta_1(I)} - G(I) \rVert}^2_2\:,
\end{equation}

where, $G(I)$ is the output of the \textit{pool5} layer of the fine tuned VGG-Face network. 
In the second step we learn the parameters of the fully connected layer, $\theta_2$ of the EMNet by training together the convolution layers, estimated in the previous step, and the fully connected layers. This step is achieved by minimizing the cross entropy loss: 
\begin{equation}
    L_2 = -\sum_{i=1}^{N} \sum_{j=1}^{M} l_{i,j} log \hat{l}_{i,j}\:,
\end{equation}
where, $l_{i,j}$ is the ground truth label and $\hat{l}_{i,j}$ is the predicted label. 

\subsection{Learning a spatio-temporal representation}
\label{sec:spatio_temporal}
For learning a spatio-temporal representation of the facial video sequences we aggregate features computed by the EMNet along trajectories generated by facial geometries (we will name it TPF-FGT from Trajectory Pooled Features from Facial Geometry Trajectories). First we detect facial geometries in a form of a fixed set of fiducial points in the whole video sequence in a per-frame fashion. To compute the fiducial points we first frontalize all the cropped face with \cite{hassner2015effective}. Then on this cropped frontalized faces we estimate the facial geometry with the with the facial alignment method \cite{kazemi2014one}. This will output $68$ fiducial landmark points on each image. The detected fiducial points are tracked across the sequence to form trajectories corresponding to specific locations on the face (e.g corners of the eyes, mouth, see Fig. \ref{fig:method} for an example). We pool features along these trajectories from the EMNet feature space. Such a pooling is advantageous because it captures the temporal relations between the frames. After reducing the dimensionality of the pooled features we learn a set of clusters over the distribution of the features using Gaussian Mixture Models (GMMs). Once the clusters are learned we use Fisher Vector (FV) \cite{sanchez2013image} encoding to produce a compact feature vector for each sequence. The final vectors are used to train a linear classifier. In the rest of section we detail the main steps of the proposed method.

\subsubsection{Trajectory pooled features} Given a sequence of images we can compute all corresponding facial geometries with the method previously presented. As each geometry is described by a fixed set of ordered points we can track these points along all the sequence to form trajectories. Along these trajectories we pool features from a feature space of choice. In our case, we use features computed at different layers of an EMNet. 

\subsubsection{Fisher Vectors} The next step is to get a single vector representation of each emotion video. On this vector an SVM classifier is trained. We choose the Fisher Vector representation for this encoding \cite{FisherJaakkola}. Each TPF is an observation vector corresponding to each landmark trajectories. We denote all the observed TPFs in the training set as $\textbf{X}$. We assume the trajectory pooled features (TPF) are drawn from a Gaussian Mixture Model (GMM). A $K$ component GMM is computed over the training set of TPF . Assuming that the observations in $\textbf{X}$ are statistically independent the log-likelihood of $\textbf{X}$ given $\vv{\theta}$ is:  
\begin{equation}
\log P(\textbf{X}\vert\vv{\theta})=\sum_{m=1}^{M} \log \sum_{k=1}^{K} w_{k} \mathcal{N}(\vv{x}_m;\vv{\mu}_{k},(\vv{\sigma}_{k})^{2})\:,
\end{equation}
where $\sum_{k=1}^{K} w_{k}=1$ and $\vv{\theta}=\{ w_{k},\vv{\mu}_{k},(\vv{\sigma}_{k})^{2}\}$. We assume diagonal covariance matrices. The parameters of the per-class GMMs are estimated with the Expectation maximization (EM) algorithm to optimize the maximum likelihood (ML) criterion. To keep the magnitude of the Fisher vector independent of the number of observations in $\textbf{X}$ we normalize it by $M$. Now we can write the closed form formulas for the gradients of the log-likelihood $P(\textbf{X} \vert \vv{\theta})$ w.r.t to the individual parameters of the GMM as:
\begin{equation}
\vv{\mathcal{J}}^{\textbf{X}}_{w_{k}}=\frac{1}{M\sqrt{w_{k}}}\sum_{m=1}^{M}\gamma_{k}(m)-w_{k}
\end{equation}
\begin{equation}
\vv{\mathcal{J}}^{\textbf{X}}_{\vv{\mu}_{k}}=\frac{1}{M \sqrt{w_{k}}}\sum_{m=1}^{M}\gamma_{k}(m)\Bigg(\frac{\vv{x}_m-\vv{\mu}_{k}}{(\vv{\sigma}_{k})^{2}}\Bigg)
\end{equation}
\begin{equation}
\vv{\mathcal{J}}^\textbf{{X}}_{(\vv{\sigma}_{k})^{2}}=\frac{1}{M\sqrt{2w_{k}}}\sum_{m=1}^{M}\gamma_{k}(m)\Bigg[\frac{(\vv{x}_m-\vv{\mu}_{k})^{2}}{(\vv{\sigma}_{k})^{2}} -1\Bigg]\:,
\end{equation}
where $\gamma_{k}(m)$ is the posterior probability or the responsibility of assigning the observation $\vv{x}_m$ to component $k$.\\
Now the FV for each video is constructed by stacking together the derivatives computed w.r.t to the components of the GMM in a single vector. The details of all the close formed formulas can be found in the following paper \cite{JakobFisher}.

\section{Experimental Results and Discussions}
\label{sec:results}
The experimental results have been conducted on the introduced \emph{SASE-FE} dataset. For comparison, we have replicated experiments on the \emph{Extended Cohn Kanade} (CK+) \cite{lucey2010extended} dataset and the Oulu-CASIA dataset \cite{zhao2011facial} and for spontaneous expression recognition we provide results of the BP4D-Spontaneous dataset \cite{ZHANGBP4D}. \\
Due to its relatively small size and simplicity, the CK+ is one of the most popular benchmarking datasets in the field of facial expression analysis. It contains 327 sequences capturing frontal poses of 118 different subjects while performing facial expressions in a controlled environment. The facial expressions are acted. Subjects' ages range between 18 and 50 years old, consisting of 69\% females and having relative ethnic diversity. Labels of presence of universal facial expressions and the Facial Action Units are provided. \\
The Oulu-CASIA dataset provides facial expressions of primary emotions in three different illumination scenarios. It includes 80 subjects between 23 to 58 years old from whom 73.8\% are males. Following other works \cite{ding2017facenet2expnet}, we only use the strong illumination partition of the data which consists of 480 video sequences (6 videos per subject). It has higher variation and constitutes a good complement to the CK+ for cross validating our method. We also test our method on the $12$ action unit recognition problem of in the BP4D-Spontaneous dataset. In this dataset, there are $41$ adults with $8$ videos each giving a total of $328$ videos. Each frame is annotated with $12$ facial AUs. In contrast with all previous set-ups, recognizing AUs is a multi-label classification problem.

In the following sections we first discuss the implementation details of each step of the proposed methodology followed by discussion of the experimental results. 

\subsection{Implementation Details}
\label{sec:implementation}

\begin{figure}
    \centering
    \includegraphics[width=0.9\linewidth, height=0.20\textwidth]{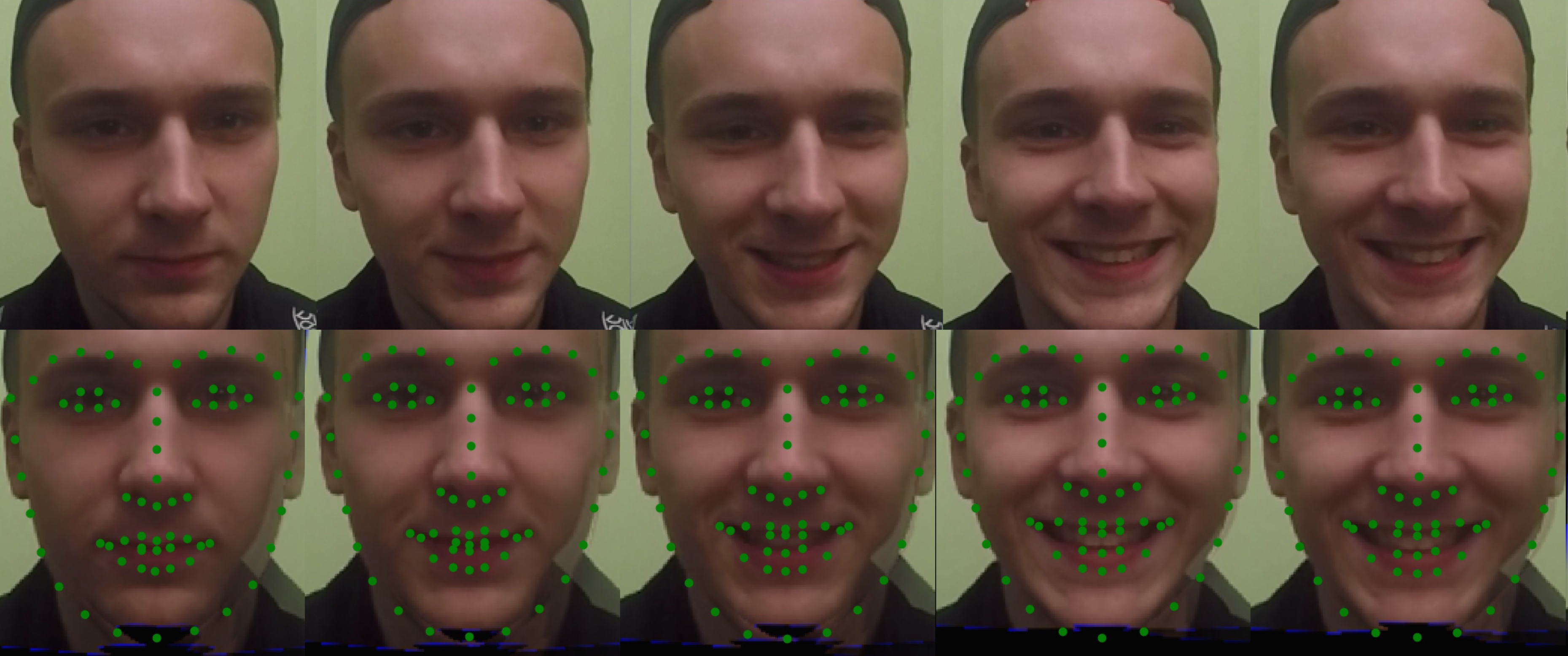}
    \caption{Illustration of the pre-processing we perform on the data. Detected faces are first extracted and frontalized and facial landmarks localised for each image in the input sequences.}
    \label{fig:alignment}
\end{figure}

The proposed methodology consists of the following steps: first, given a video sequence we extract faces from background, frontalize them and localize facial landmarks (see Fig. \ref{fig:alignment}). Second, we fine-tune a pretrained VGG-Face deep network \cite{Parkhi15} for recognising facial expressions. Third, we use this network for guiding the training of a so called EMNet following work proposed in \cite{ding2017facenet2expnet} (see also Sec. \ref{sec:knowledge_transfer}). This second network is used to compute static representations from still images. Fourth, we pool features from the previously computed static representation space along trajectories determined by the facial landmarks. Fifth, we compute fixed length descriptors for each video sequence using the Fisher Vector encoding. These final descriptors are then classified with a linear SVM. We use a leave-one-actor-out validation framework for all our experiments. For the theoretical framework of the spatio-temporal representation and the knowledge transfer training approach of the EMNet, please refer to Sec. \ref{sec:method}. For a visual overview of the method see Fig. \ref{fig:method}.

\textbf{Preprocessing}. We first extract faces from the video sequences. After faces are extracted we perform a frontalization which registers faces to a reference frontal face by using the method of Hassner et al. \cite{hassner2015effective}. This removes variance in the data caused by rotations and scaling. This frontalization method estimates a projection matrix between a set of detected points on the input face and a reference face. This is then used to back-project input intensities to the reference coordinate system. Self-occluded regions are completed in an aesthetically pleasant way by using color information of the neighbouring visible regions and symmetry. Finally in all synthesised frontal faces we estimated the facial geometry, using a classical, robust facial alignment method \cite{kazemi2014one} trained to find 68 points on the image (an example of the frontalization process is showed in Fig. \ref{fig:alignment}).   

\textbf{Fine-Tuning the VGG-Face}. For all experiments, including fine tuning of the VGG-FACE are done in a 10-fold cross validation for the CK+ and Oulu-CASIA datasets to keep the experiments consistent with \cite{ding2017facenet2expnet}. We define a train set of $40$ actors, validation set of $5$ actors and a test set of $5$ actors for the SASE-FE dataset. This set is exactly similar to the partitions defined in \cite{wan2017results}. Here we estimate the parameters of our proposed method on the validation set and final results are reported on the unseen test set. Here we also perform an additional experiment, since the training data is limited, we augment the training set of the SASE-FE dataset with additional training data from the Oulu-CASIA \cite{zhao2011facial} and CK+ datasets. These experiments are denoted as \textit{Data Augmentation}. The training is done for 200 epochs with a learning rate of $0.001$. It is decreased every 50 epochs. The fully connected layers are randomly initialised with the Gaussian distribution. The min-batch size is $32$ and the momentum is $0.9$. The dropout is set to $0.5$. From each frame the face is cropped and scaled to $224\times224$. The bottom two convolution layers are left unchanged. In the testing phase, if the CNN is able to recognise more than $50\%$ of the  frames in the video correctly then the video is deemed to be correctly classified. For the $6$ genuine class and the $6$ unfelt class experiment the network is trained for the $12$ class problem, and the final fully connected layer is retrained with the appropriate number of classes.  

\textbf{Training the EMNet}. The architecture of EMNet is the same as the one proposed in \cite{ding2017facenet2expnet}. It consists of $5$ convolutional layers each followed by a ReLU activation and a max pooling layer. The filter size of the convolutions layers is $3\times3$ and that of the pooling layer is $3\times3$ with a stride of $2$. The output of each layer is $64,128,256,512,512$. Furthermore, we need to add another $1\times1$ convolutional layer to match the dimensionality of the output of the EMNet to the $pool5$ layer of the fine tuned VGG-Face net for the regularisation in the first step. We append a single fully connected layer of size $256$. We just use one layer to prevent overfitting. We use this size of $256$ for distinguishing between all multi-class experiments of classifying all emotions in the dataset. The size of the fully connected layer is further reduced to $128$ for the binary classification experiment of distinguishing between genuine and unfelt FEEs. This is because the training data available for binary classification is much less than the training data for classifying all emotion.

\textbf{Trajectory pooled features (TPF)}. The TPFs from the facial geometry trajectories (TPF-FGT) are aggregated in a rectangular region of pixel size $64\times64$ which we have experimentally set. This size is scaled by a ratio of the size of the input image and the feature map from the corresponding layer of the neural network. For our experiments we use the TPF descriptors extracted from the conv5 of the EMNet. In order to train the Fisher vector for encoding we perform PCA to decorrelate the dimensions.  We experimentally set the number of first principal components to $32$. 

\textbf{Fisher Vectors encoding and classification}. For encoding the TPFs into lower dimensional representations we used the Fisher Vector encoding. Its efficacy for video analysis  has been proven for action recognition \cite{OneataFisher}. In order to train GMMs, we first decorrelate the dimensions of the TPFs with PCA and reduce its dimension to $d$. Then, we train a GMM with $k=16$ mixtures. We can use a low value for $k$ as compared to other papers in the literature because the trajectory computed on the landmarks is already discriminative as compared to the dense trajectory features. This enables us to construct a compact feature representation with FV which is also discriminative. Moreover, we square-root normalise followed by the $L2$ norm of each vector. The video is represented with a $2kd$ dimensional vector. We use the Fisher Vectors to train  a linear SVM for classification. The value of the regularisation parameter is set to $C=100$. The parameters $K$ and $C$ were set using the validation set and then tested on the unknown test set of the SASE-FE dataset.

\begin{table}
\centering
\caption{Our method shows state-of-the-art results when compared with best performing setups on the CK+ dataset. This proves generalisation capacity of this approach.}
 \begin{tabular}{|c|c|} 
 \hline
 \textbf{Method} & \textbf{Accuracy(\%)} \\ [0.5ex] 
 \hline\hline
 AURF \cite{LiuAURF} & 92.22 \\ \hline
 AUDN\cite{LiuAUDN} & 93.70 \\ \hline
 STM-Explet\cite{STM-Expletliu2014learning} &  94.2 \\\hline
 LOmo \cite{Lomosikka}& 95.1  \\\hline
 IDT+FV \cite{idtEmotionsAfshar} &  95.80\\ \hline
 Deep Belief Network \cite{LiuDBN} &  96.70 \\ \hline 
 Zero-Bias-CNN \cite{Zero-bias-CNN} & 98.4 \\\hline
 Ours-Final & \textbf{98.7} \\\hline
\end{tabular}
\label{tab:perf_ck}
\end{table}

\begin{table}
\centering
\caption{Our method shows state-of-the-art results when compared with best performing setups on the Oulu-CASIA dataset. This proves the generalization capacity of such an approach.}
 \begin{tabular}{|c|c|} 
 \hline
 \textbf{Method} & \textbf{Accuracy (\%)} \\ [0.5ex] 
 \hline
 DTAGN \cite{DTAGN} & 81.46 \\\hline
 LOmo \cite{Lomosikka}& 82.10  \\\hline
 PPDN \cite{zhao2016peak} &  84.59\\ \hline
 FN2EN \cite{ding2017facenet2expnet}&  87.71 \\\hline
 Ours-Final & \textbf{89.60} \\\hline
 \end{tabular}
 \label{tab:sota_Oulu}
\end{table}

\subsection{Discussion}
In this section, we discuss the experimental results obtained by our proposed method. For brevity, we have denoted both in the text and figures the genuine FEE labels by adding a \emph{G} in front of the labels (e.g GSad) and the corresponding unfelt FEE by adding an \emph{U} in the same fashion (e.g UAnger). We start by discussing results on the \emph{Cohn-Kanade}, the Oulu-CASIA and BP4D-Spontaneous datasets and then we discuss the results on the proposed SASE-FE dataset. 

\subsubsection{CK+} The performance of several state-of-the-art methods and the performance of our final method is given in Table \ref{tab:perf_ck}. We are able to come very close to the state of the art performance on this dataset.\\
In terms of methodology, \cite{idtEmotionsAfshar} is the closest method to our proposed method. The authors of this paper implement the improved dense trajectories framework proposed for action recognition \cite{wang2013dense} for emotion recognition. We are able to improve their results by aggregating the feature maps along the fiducial points and computing the TPF-FGT features. \\
We observe that our method is better than methods which use a per frame feature representation rather than per-video as in our case \cite{Lomosikka,STM-Expletliu2014learning}. In \cite{Lomosikka}, this per-frame feature is the concatenation of SIFT features computed around landmark points, head pose and local binary patterns (LBP). They propose a weakly supervised classifier which learns the events which define the emotion as hidden variables. The classifier is a support vector machine which was estimated using the multiple-kernel learning method. From the table we can observe that when landmarks are used along with the CNN feature maps we are able to top their performance. The rest of the methods listed in the table use deep learning techniques to classify emotions \cite{LiuDBN,LiuAURF,Zero-bias-CNN}. They design networks able to specifically learn facial AUs. We can observe that we out perform the best performing method \cite{Zero-bias-CNN} on the CK+ dataset. 

\subsubsection{Oulu-CASIA}
We also , show the efficacy of our method on a more difficult dataset like the Oulu-CASIA dataset. In Table \ref{tab:sota_Oulu} we can observe that our method outperforms the previous best performance of \cite{ding2017facenet2expnet} by $1.9\%$. In Table \ref{tab:emotionwise} we show the emotion-wise comparison between our proposed method and \cite{ding2017facenet2expnet}. The two main differences between  \cite{ding2017facenet2expnet} and our method are that we align the faces and then add the TPFs for classification. In our experiments we observed that aligning the faces on the Oulu-CASIA dataset gave only very marginal improvement while once we add the TPFs for classification then we can get significant improvements. The improvements are especially observed in three emotions Anger, Disgust and Sadness. These emotions are typically confused between each other. This experiment shows that the temporal information is important for emotion recognition. 

\begin{table}
\centering
\caption{Emotion-wise comparison between our proposed method and \cite{ding2017facenet2expnet} on the Oulu-CASIA dataset.}
 \begin{tabular}{|c|c|c|} 
 \hline
 \textbf{Emotion} & \textbf{Accuracy} \cite{ding2017facenet2expnet} (\%) & \textbf{Accuracy} \textbf{[Ours-Final]} (\%)\\[0.5ex] 
 \hline
 Anger & 75.2 & 80.1 \\ \hline
 Disgust & 87.3 & 88.0 \\ \hline
 Fear & 94.9 & 95.1 \\ \hline
 Happiness & 90.8& 89.7 \\\hline
 Sadness & 88.4 & 91.3 \\\hline
 Surprise & 92.0 & 92.7  \\\hline\hline
 Average & 87.7 & 89.6 \\ \hline
 \end{tabular}
  \label{tab:emotionwise}
\end{table}

\subsubsection{BP4D-Spontaneous}
Considerably more challenging is the recognition of spontaneous expression of emotion. For this purpose we show results on the BP4D dataset. The evaluation is done in the 3-fold cross validation framework. The evaluation metrics is F1-segment score which is the harmonic mean of the precision and recall. We do the following steps to achieve the final results. First, we finetune the VGG-FACE network on the 12 action units. We sample 100 frames as positive and 200 frames as negative examples per sequence as done in \cite{DRML}. Then we train the EMnet from VGG-FACE network to do AU recognition. From the EMnet we compute the TPF and then finally the SVM for classification of AUs. We compute a F1-segment score as opposed to F1-frame score as done in \cite{DRML} because the trajectories on the landmark-points are computed over a $16$ frame symmetric window around each frame. For each video in the dataset the first and the last $8$ frames were discarded. We found that this window size was a good choice. If a large window was used then the Fisher vectors which are constructed for the segments are not discriminative. 

The results of comparison of our framework with the state of the art are presented in Table \ref{BP4D}. As we can see the method trained to recognise a single emotion label does not perform competitively as compared to the state-of-the-art. This is because the methods which are designed to do AU recognition are trained via local patches as opposed to the trajectories from all the face landmarks. Since we know the location of the action units we automatically selected the trajectories to train the final SVM. For example if the AU is a lip corner depressor we choose the trajectories from the patch where the action unit is most likely to occur. We know this location because of the landmark points. This result is represented as $Ours-Final+SF$ in table \ref{BP4D}. Additionally AUs can co-occur. Therefore, we weight the final recognition scores of the SVM with the co-occurrence probability of the AU. We estimate this probability matrix from the training data. This result is shown as \textit{Proposed+ SF+ CO} in table \ref{BP4D}. This way we can show that our method is competitive for dynamic spontaneous AU recognition. If one explicitly estimates the spatial representation temporal modelling and AU correlation then this method can achieve a higher accuracy. This is done with a CNN and LSTM in \cite{CNNLSTMAU}.

\begin{table}[t]
\centering
\caption{This table presents the comparison of our method with the state-of-the-art on the BP4D dataset.}
\label{BP4D}
\begin{tabular}{|l|c|}
\hline
\textbf{Method} & \multicolumn{1}{c|}{\begin{tabular}[c]{@{}c@{}} \textbf{Average} \\ \textbf{F1-score} \end{tabular}} \\\hline
LSVM-HOG \cite{DRML} & 32.5    \\\hline
JPML \cite{JPML}& 45.9  \\\hline
AlexNet \cite{DRML}   & 38.4 \\\hline
Ours-Final  & 43.6  \\\hline
Ours-Final + SF & 46.8  \\\hline
Ours-Final + SF + CO & 48.1 \\\hline
DRML \cite{DRML}&  48.3 \\\hline
CNN + LSTM \cite{CNNLSTMAU} & \textbf{53.9} \\\hline
\end{tabular}
\end{table}

\subsubsection{SASE-FE} 
\begin{figure}[h]
    \centering
    \subfigure[]
    {
        \includegraphics[width=0.8\linewidth]{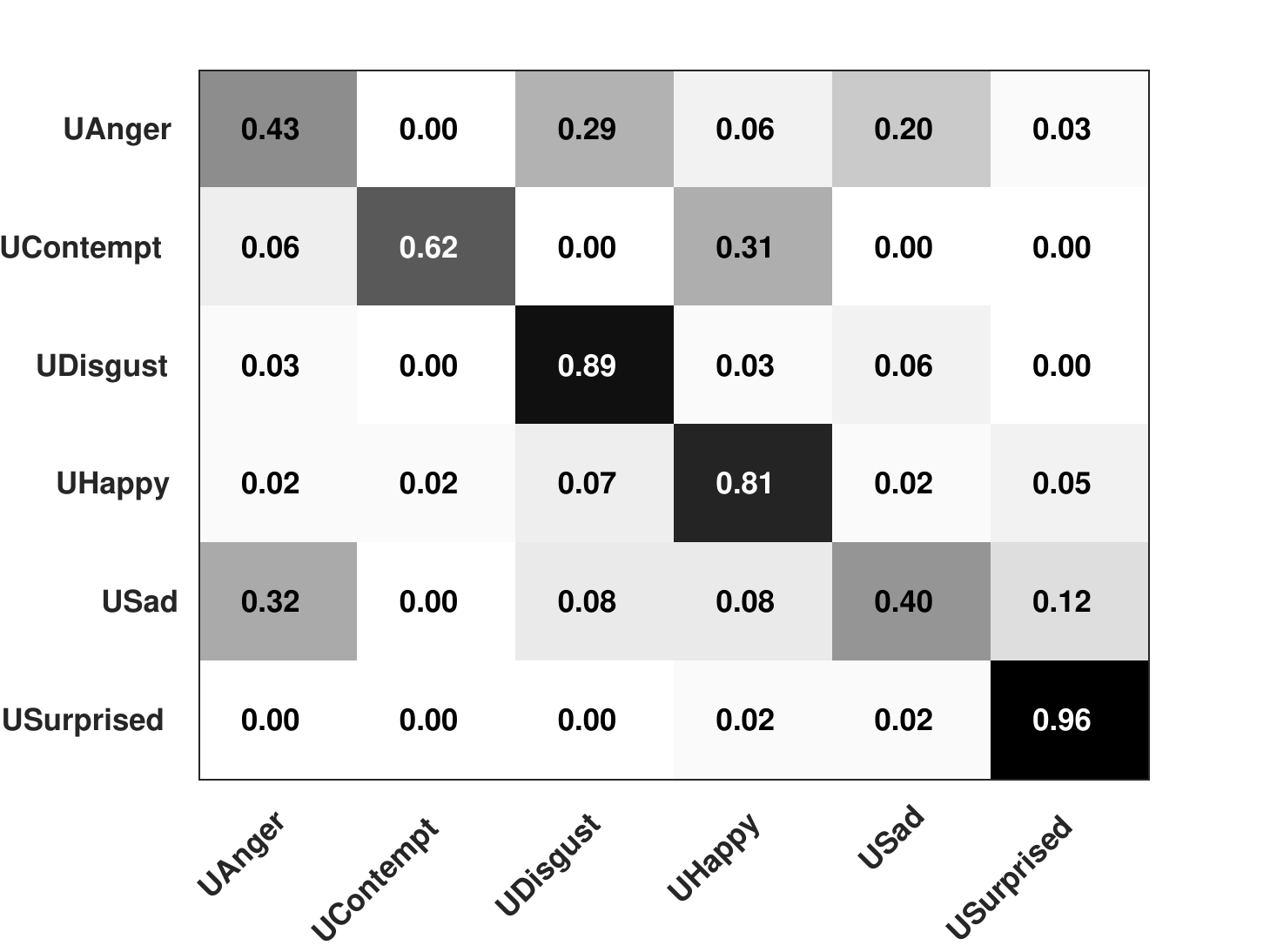}
        \label{fig:confmat_6fake}
    }\\
    \subfigure[]
    {
        \includegraphics[width=0.8\linewidth]{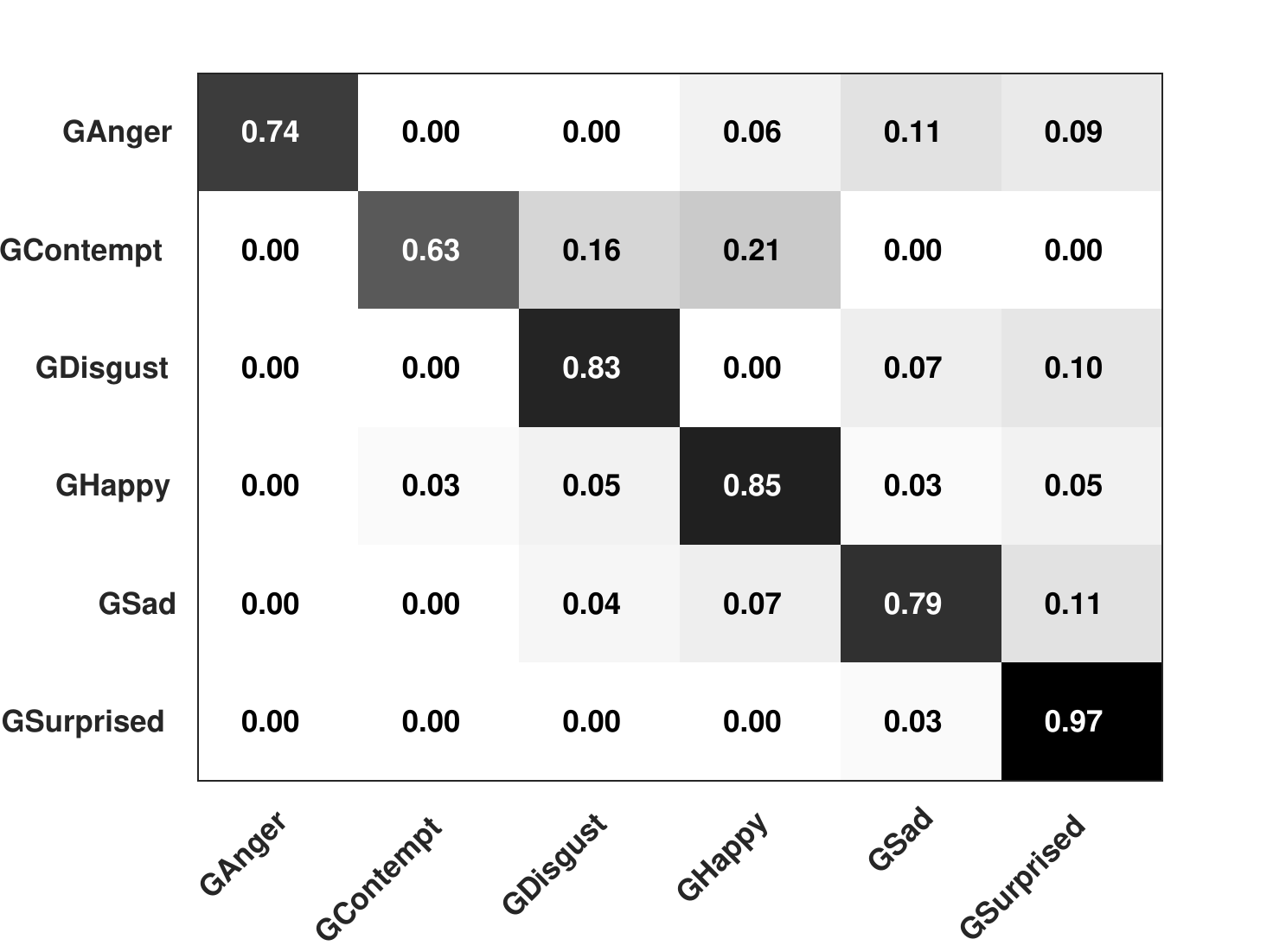}
        \label{fig:confmat_6true}
    }
    \caption{Confusion matrices for 6 classes classification. (a) 6 class classification on the unfelt subset of SASE-FE. (b) 6 class classification on the genuine subset of SASE-FE. Genuine FEEs are labelled with an initial 'G'  and unfelt FEEs with an 'U'.}
    \label{fig:confmat_6}
\end{figure}

The set of experiments we present in this section has been designed with the purpose of exploring spatial and temporal representation for the proposed problem. We will show how results improve  by increased use of domain knowledge for encoding temporal information and by using specially learned representations. Furthermore, we can see more improvement in the recognition results from learning a EMNet from a finetuned VGG-Facenet. For example, in the first conducted experiment we globally extract a handcrafted descriptor (SIFT) and we disregard any temporal information. On the proposed dataset, this produces results slightly above chance. By computing local descriptors around Improved Dense Trajectories (IDT), a proven technique in the action recognition literature, we obtain a small improvement. While the tracked trajectories follow salient points, there is no guarantee that these points are fiducial points on the face. Because fiducial points are semantically representative on the facial geometry, they are usually best for capturing local variations due to changes of expression. This assumption is confirmed by extracting local descriptors around landmark trajectories produced by the facial geometry detector. In the final setup, the best performance is obtained by extracting the representation from a feature space produced by the EMNet CNN. In Table \ref{tab:perf_sasefe} we compare the performance between the TPF-FGT obtained from the last convolution layer of both the VGG-Face and EMNet. Since the EMNet is trained only for the emotion recognition domain the performance of the EMNet is higher than that of the VGG-Face.\\
In terms of the use of temporal information several comments can be made. In line with the literature, temporal information is essential in improving recognition of subtle facial expressions. What we are presenting is by no chance an exhaustive study. While a state-of-the-art method in producing compact representations of videos, Fisher Vectors encoding disregards some of the temporal information for compactness. Other, more powerful sequential learning methods, like Recurrent Neural Networks, might be employed with better results.\\
In Fig. \ref{fig:confmat_6} we present confusion matrices for a six class classification problem on the proposed dataset. We split the classification problem in two, training on the 6 genuine and the 6 unfelt emotions respectively. On the SASE-FE, several observations can be made. Both in the case of genuine and unfelt FEE classifications, the expressions that are easier to discriminate are Happiness and Surprise. This due to their particularly distinctive morphological patterns. The most difficult expression to distinguish is contempt, which is in alignment with the literature and with the result on the CK+, the benchmark dataset as previously explained. On average, the proposed method gets better results when trying to discriminate between the genuine emotions than when discriminating between the unfelt ones. This is to be expected, taking into account that when faking the expressions, the subjects are trying to hide a different emotional state. This will introduce particular morphological and dynamical changes that makes the problem more difficult. Particularly interesting is the difficulty the classifier has in recognizing unfelt sadness. The high level of confusion with unfelt anger should be noticed along with the fact that this is not the case for genuine emotions.\\

\begin{figure}
    \includegraphics[width=0.8\linewidth]{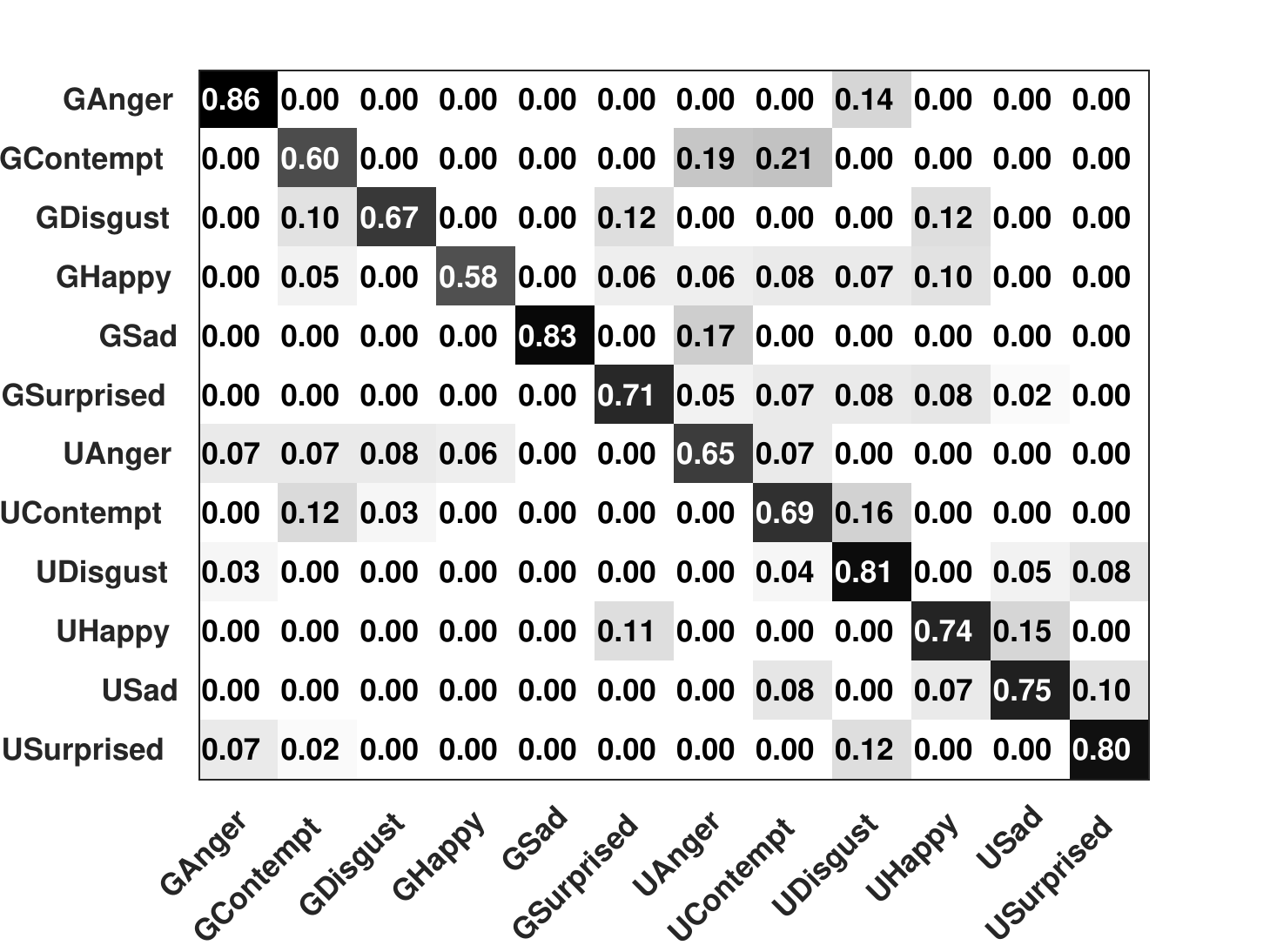}
    \caption{Confusion matrix for 12 class classification on the SASE-FE dataset. Genuine FEE are labelled with an initial 'G'  and unfelt FEE with an 'U' .}
    \label{fig:confmat_12}
\end{figure}
In Fig. \ref{fig:confmat_12} we present the confusion matrix for the problem of classifying between all 12 classes (genuine and unfelt jointly). This can be interpreted together with results in Table \ref{tab:perf_sasefe} where we present classification accuracies for each pair (genuine/unfelt). When trained with all classes, the best results are obtained for genuine sadness and the worst for genuine contempt and genuine contempt. In Table \ref{t4}, overall accuracies of especially the unfelt ones remain low, which underlines again the difficulty of the problem and suggests more powerful sequential learning tools should be employed. Interestingly, it is easiest to discriminate between genuine and unfelt expressions of anger which is due to the fact that anger is recognised a lot by the activation of muscles in the eye region. Also the results show that the recognition rate of the unfelt expressed contempt is by chance, i.e. contempt is easier to unfelt, hence more difficult to detect, and this is due to the fact that the main facial features expressing this emotion are mainly around the mouth region which can be quickly and easily moved, whereas muscles around the eyes (which are important in expressing other emotions) are not instantly deformable by signals from brain.\\
Table \ref{iccv_tab} shows the comparison of the average recognition rate for a 12-class classification between recently proposed techniques reported in \cite{wan2017results} and the proposed method. These results correspond to the winning methods from the ChaLearn international competition we organize at ICCV 2017. We outperform these winning methods. In this table, we can also observe that our proposed method outperforms the LSTM based approaches \cite{tani2004self}. This is because in the temporal stage we used a hand tuned approach which requires fewer parameters to be tuned as compared to a LSTM. This advantage would be negated on a very large datasets but nevertheless it demonstrates the efficiency of our method.

\begin{table}[h]
\centering
\caption{Genuine vs unfelt FEE classification performance on the SASE-FE dataset.}
 \begin{tabular}{|c|c|c|} 
 \hline
\textbf{Emotion Pair} & \textbf{Accuracy Genuine (\%)} & \textbf{Accuracy Unfelt (\%)} \\ \hline  
Anger & 72.5 & 66.3 \\  \hline
Happiness & 76.7 & 65.4 \\  \hline
Sadness & 71.5 & 61.3 \\\hline
Disgust & 66.4 & 59.7 \\\hline
Contempt & 63.4 & 58.3 \\\hline
Surprise & 71.3 & 63.4 \\\hline
\end{tabular}
\label{t4}
\end{table}

\begin{table}[h]
\centering
\caption{The average recognition rate for 12 class classification between several state-of-the-art methods \cite{wan2017results} and the proposed method; DA=Data augmentation.}
\label{iccv_tab}
\begin{tabular}{|c|c|}
\hline
\textbf{Method} & \multicolumn{1}{c|}{\begin{tabular}[c]{@{}c@{}}Accuracy \end{tabular}} \\\hline \hline
 Rank-SVM \cite{joachims2002optimizing}       & 66.67   \\\hline
LSTM-PB \cite{tani2004self}       & 66.67  \\\hline
CBP-SVM \cite{gao2016compact}       & 65.00  \\\hline
HOG-LSTM \cite{pei2017temporal}      & 61.70 \\\hline
CNN   \cite{mallya2016learning}     & 51.70  \\\hline
Ours-Final & 68.7  \\\hline
Ours-Final + DA & \textbf{70.2} \\\hline
\end{tabular}
\end{table}

\begin{table}[h]
\centering
\caption{Performance on the SASE-FE dataset. IDT = Improved dense Trajectories, FGT= Facial Geometry Trajectories, TPF-IDT = Trajectory Pooled Features along IDT, TPF-FGT = Trajectory Pooled Features along FGT, DA = Data Augmentation, $^1$ Fine-tune, no data augment, $^2$ Fine-tune, data augment.}
\resizebox{0.5\textwidth}{!}{ 
 \begin{tabular}{|c|c|c|} 
 \hline
 & {Method} & {Accuracy(\%)} \\ [0.5ex] \hline
 \multirow{8}{*}{{12 classes}} & SIFT+FV & 12.2 \\\cline{2-3}
 & TPF-FGT(SIFT)+TPF-IDT(MBH)+FV & 21.3    \\\cline{2-3}
 & VGG-Face$^1$ & 39.5\\\cline{2-3}
 & VGG-Face$^2$ & 49.8\\\cline{2-3}
 & TPF-FGT(VGG-Face)+FV & 50.2\\\cline{2-3} 
 & TPF-FGT(VGG-Face)+FV+Aligned Faces & 54.3 \\\cline{2-3}
 & TPF-FGT(VGG-Face)+FV+Aligned Faces+DA & 60.3\\\cline{2-3}
 & TPF-FGT(EMNet)+FV & 65.7\\\cline{2-3} 
 & TPF-FGT(EMNet)+FV Aligned Faces & 68.7\\\cline{2-3}
 & {TPF-FGT(EMNet)+FV Aligned Faces+DA} & {70.2}\\\hline \hline

 \multirow{7}{*}{{6 classes (genuine)}} & VGG$^1$ & 65.2\\\cline{2-3}
 & VGG $^2$ & 71.7\\\cline{2-3}
 & TPF-FGT(VGG-Face)+FV & 73.7\\\cline{2-3} 
 & TPF-FGT(VGG-Face)+FV Aligned Faces & 74.2 \\\cline{2-3}
 & TPF-FGT(VGG-Face) +FV Aligned Faces+ DA & 76.5\\\cline{2-3}
 & TPF-FGT(EMNet) +FV & 77.2\\\cline{2-3} 
 & TPF-FGT(EMNet) +FV Aligned Faces & 78.7\\\cline{2-3}
 & {TPF-FGT(EMNet) +FV Aligned Faces+ DA} & {80.3}\\\hline \hline
 
 \multirow{7}{*}{{6 classes (unfelt)}} & VGG$^1$ & 42.7\\\cline{2-3}
 & VGG$^2$ & 59.2\\\cline{2-3}
 & TPF-FGT(VGG-Face) + FV & 62.3\\\cline{2-3}
 & TPF-FGT(VGG-Face) + FV + Aligned Faces & 64.2 \\\cline{2-3}
 & TPF-FGT(VGG-Face) + FV + Aligned Faces + DA & 67.5\\\cline{2-3}
 & TPF-FGT(EMNet) + FV & 70.3\\\cline{2-3} 
 & TPF-FGT(EMNet) + FV + Aligned Faces & 72.2\\\cline{2-3}
 & {TPF-FGT(EMNet)+FV Aligned Faces+DA (Ours-Final)} & {73.6}\\\hline
\end{tabular}
}
\label{tab:perf_sasefe}
\end{table}

\section{Conclusion}
\label{sec:conclusion}
Previous research from psychology suggests that discriminating the genuineness of feelings or intentions hidden behind facial expressions is not a well mastered skill. For this reason, we provide for the first time a dataset capturing humans while expressing genuine and unfelt facial expressions of emotion at high resolution and a high frame rate.\\  
In this paper, we also propose a method inspired from action recognition and extend it to perform facial expression of emotion recognition. We combine the feature maps computed from the EMNet CNN with a facial landmark detector to compute spatio-temporal TPF descriptors. We encode these descriptors with Fisher vectors to get a single vector representation per video. The feature vector per video is used to train a linear SVM classifier. We outperform the state of the art performance on the the publicly available CK+ and Oulu-CASIA both containing posed FEEs, and show competitive results on the BP4D dataset for facial action unit recognition. Furthermore, we provide several baselines on our SASE-FE dataset. We also improve the results of the winning solutions of the recent ChaLearn competition about our dataset. We show that even though we obtain good results on the $6$ class genuine and unfelt problem, the $12$ class and the binary emotion pair classification problem still remains a challenge. This is because the distinguishing factors between the unfelt and genuine expressions occur in a very short part of the whole emotion and are a challenge to model. \\
This preliminary analysis opens several future lines of research. Our experiments showed two most important problems of current state of the art methods. Firstly, current state of the art CNNs, such as VGG-Face, do not work at the required spatial resolution to detect minute changes in facial muscle movements, which are required to differentiate and distinguish between unfelt FEEs.  Secondly, alternative temporal analysis strategies could be considered to analyse SASE-FE at high fps, which may include variants of Recurrent Neural Nets or 3D-CNNs approaches.

\ifCLASSOPTIONcompsoc
  \section*{Acknowledgments}
\else
  \section*{Acknowledgment}
\fi

This work is supported Estonian Research Council Grant (PUT638), the Estonian Centre of Excellence in IT (EXCITE) funded by the European Regional Development Fund, the Spanish Project TIN2016-74946-P (MINECO/FEDER, UE) and CERCA Programme / Generalitat de Catalunya. This project has received funding from the European Union’s Horizon 2020 research and innovation programme under the Marie Sklodowska-Curie grant agreement Nº 665919. We gratefully acknowledge the support of NVIDIA Corporation with the donation of the Titan Xp GPU used for this research.

\ifCLASSOPTIONcaptionsoff
  \newpage
\fi
\bibliographystyle{IEEEtran}
\bibliography{ms}

\begin{IEEEbiography}[{\includegraphics[width=1in,height=1.2in,clip]{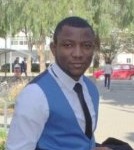}}]{Ikechukwu Ofodile} obtained his BSc in Electrical and Electronics Engineering from Eastern Mediterranean University, North Cyprus. He is currently a MSc Student and a member of the intelligent computer vision (iCV) research group at the University of Tartu, Estonia. He is also a member of Philosopher, the Estonian Robocup team of the University of Tartu and a member of the Estonian student satellite project working on attitude determination and control of ESTCube-2. His research interests include machine learning, pattern recognition and HCI as well as control engineering and attitude control system design for nanosatellites and microsatellites.
\end{IEEEbiography}
\vskip -2\baselineskip plus -1fil

\begin{IEEEbiography}[{\includegraphics[width=1in,height=1.0in,clip]{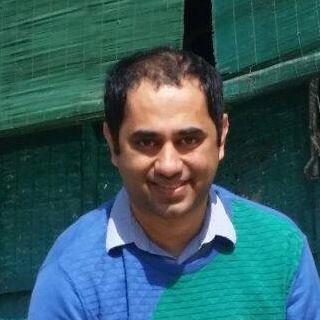}}]{Kaustubh Kulkarni}
obtained in Bachelors in engineering from Mumbai university. He completed his MSc. from Auburn University, USA. Following which he worked at Siemens research labs in India and USA. He is in the process of getting his PhD from INRIA, Grenoble, France. Currently, he is working at the Computer Vision Center at Universitat Autònoma de Barcelona. He has experience working in medical image analysis, action recognition, speech recognition and emotion recognition. 
\end{IEEEbiography}
\vskip -2\baselineskip plus -1fil

\begin{IEEEbiography}
[{\includegraphics[width=1in,height=1.25in,clip]{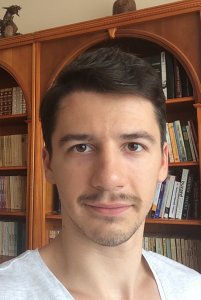}}]{Ciprian Adrian Corneanu} got his MSc in Computer Vision from Universitat Aut\'{o}noma de Barcelona in 2015. Currently he is a PhD student at the Universitat de Barcelona and a fellow of the Computer Vision Center from Universitat Aut\'{o}noma de Barcelona. His main research interests include face and behavior analysis, affective computing, social signal processing and human computer interaction.
\end{IEEEbiography}
\vskip -2\baselineskip plus -1fil

\begin{IEEEbiography}[{\includegraphics[width=1in,height=1.0in,clip]{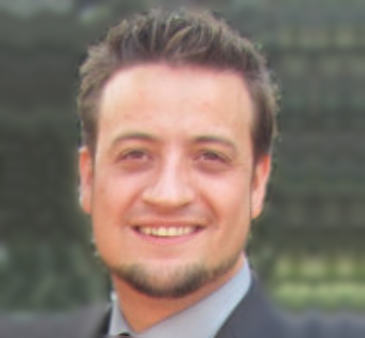}}]{Sergio Escalera} is an associate professor at the Department of Mathematics and Informatics, Universitat de Barcelona. He is an adjunct professor at Universitat Oberta de Catalunya, Aalborg University, and Dalhousie University. He obtained the PhD degree on Multi-class visual categorization systems at the Computer Vision Center, UAB. He obtained the 2008 best Thesis award on Computer Science at Universitat Autònoma de Barcelona. He leads the Human Pose Recovery and Behavior Analysis Group at UB and CVC. He is also a member of the Computer Vision Center at Campus UAB. He is an expert in human behavior analysis in temporal series, statistical pattern recognition, visual object recognition, and HCI systems, with special interest in human pose recovery and behavior analysis from multi-modal data. He is vice-president of ChaLearn Challenges in Machine Learning, leading ChaLearn Looking at People events. He is Chair of IAPR TC-12: Multimedia and visual information systems. 
\end{IEEEbiography}
\vskip -2\baselineskip plus -1fil

\begin{IEEEbiography}[{\includegraphics[width=1in,height=1.0in,clip]{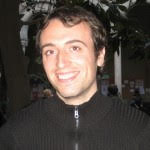}}]{Xavier Bar\'o} received his B.S. degree in Computer Science at the UAB in 2003. In 2005 he obtained his M.S. degree in Computer Science at UAB, and in 2009 the PhD degree in Computer Engineering. At the present he is associate professor and researcher at the Computer Science, Multimedia and Telecommunications department at Universitat Oberta de Catalunya (UOC). 
\end{IEEEbiography}
\vskip -2\baselineskip plus -1fil

\begin{IEEEbiography}[{\includegraphics[width=1in,height=1.0in,clip]{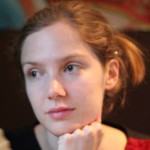}}]{Sylwia Hyniewska} received a double PhD degree from the Telecom ParisTech Institute of Science and Technology and the University of Geneva. She finished her doctoral school at the “Swiss National Center for Affective Sciences” (CISA). Afterward, she worked as an independent research Fellow of the Japan Society for the Promotion of Science (JSPS) at Kyoto University, where she collaborated with world-renowned specialists in social and emotion perception. Since 2014 she has worked at the University of Bath on topics related to emotion perception, virtual reality and pervasive devices and is a member of the Centre for Applied Autism Research (CAAR). In 2017, she joined the Institute of Physiology and Pathology of Hearing, Poland. At the institute's Bioimaging Research Center, she studies neurofeedback applied to brain fingerprinting (EEG and fMRI methods) in attentional and affective tasks.
\end{IEEEbiography}
\vskip -2\baselineskip plus -1fil

\begin{IEEEbiography}[{\includegraphics[width=1in,height=1.2in,clip]{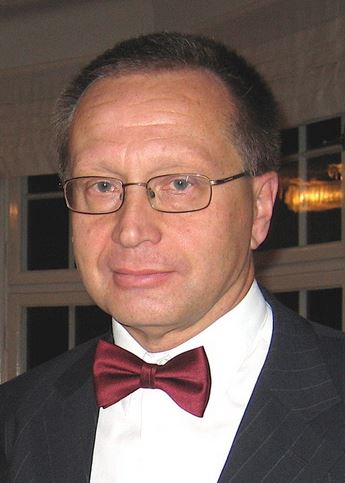}}]{J\"uri Allik}
was Candidate of Science (PhD), University of Moscow (1976) and obtained PhD in psychology from the University of Tampere, Finland (1991). He has been Chairman of Estonian Science Foundation (2003-2009), Professor of Psychophysics (1992-2002) and Professor of Experimental Psychology (2002- ) at the University of Tartu. He was also Dean of Faculty of Social Sciences (1996-2001), President (1988-1994) and Vice-President (1994-2001) of the Estonian Psychological Association. He served as a Foreign Member of the Finnish Academy of Science and Letters (1997). He has has received many awards including Estonian National Science Award in Social Sciences category (1998, 2005). He was a member of the Estonian Academy of Sciences. His research interests are psychology, perception, personality and neuroscience and his research works have received over 14,000 citations.
\end{IEEEbiography}
\vskip -2\baselineskip plus -1fil

\begin{IEEEbiography}[{\includegraphics[width=1in,height=1.05in,clip]{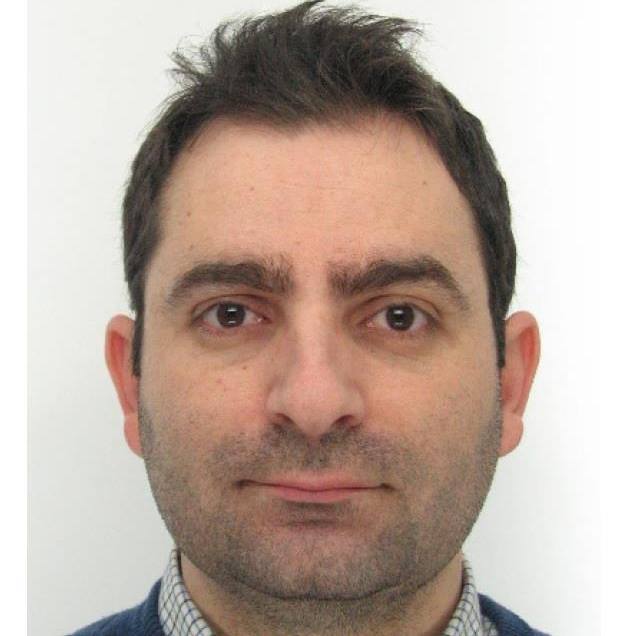}}]{Gholamreza Anbarjafari} is heading the intelligent computer vision (iCV) research group in the Institute of Technology at the University of Tartu. He is also Deputy Scientific Coordinator of the European Network on Integrating Vision and Language (iV\&L Net) ICT COST Action IC1307. He is Associate Editor and Guest Lead Editor of several journals, Special Issues and Book projects. He is an IEEE Senior member and the Vice Chair of Signal Processing/Circuits and Systems/Solid-State Circuits Joint Societies Chapter of IEEE Estonian section. He has got Estonian Research Council Grant (PUT638) and the Scientific and Technological Research Council of Turkey (T\"UBİTAK) (Proje 1001 - 116E097) in 2015 and 2016, respectively. He has been involved in many national and international industrial projects mainly related to affective computing. He is expert in computer vision, human-robot interaction, graphical models and artificial intelligence. He has been in the organizing committee and technical committee of SIU, ICOSST, ICGIP, SampTA and FG. He has been organizing challenges and workshops in FG17, CVPR17, and ICCV17. He is Associate Editor of SIVP and have organized several SI on human behaviour analysis in JIVP and MVAP.
\end{IEEEbiography}

\end{document}